\documentclass[letterpaper, 10 pt, conference]{ieeeconf}  % Comment this line out if you need a4paper

\IEEEoverridecommandlockouts                              % This command is only needed if 
\title{\LARGE \bf
\metabbr: Teaching Robot to Follow Nonverbal Human Intent 

from Egocentric Vision and Gaze
}

\usepackage{cite}
\usepackage{amsmath,amssymb,amsfonts}
\usepackage{algorithmic}
\usepackage{graphicx}
\usepackage{textcomp}
\usepackage{xcolor}
\usepackage[ruled,vlined,linesnumbered]{algorithm2e}
\def\BibTeX{{\rm B\kern-.05em{\sc i\kern-.025em b}\kern-.08em
    T\kern-.1667em\lower.7ex\hbox{E}\kern-.125emX}}
\usepackage{xspace}
\newcommand{\metabbr}{EDITH\xspace}
\usepackage{hyperref}
\definecolor{softblue}{RGB}{70,120,190}
\definecolor{forestgreen}{RGB}{34,139,34}
\hypersetup{colorlinks=true, citecolor=red, linkcolor=softblue, urlcolor=softblue}

\AtBeginDocument{\let\oldref\ref \renewcommand{\ref}[1]{\textcolor{softblue}{\oldref{#1}}}}

\begin{document}
% EDITH: a framework to teach robot to see what you see from smart glasses

%\author{Anonymous Author(s)}
\author{Dongjun Lee\textsuperscript{1*}, Juheon Choi\textsuperscript{1*}, Dong Kyu Shin\textsuperscript{1*}, Sinjae Kang\textsuperscript{1}, and Kimin Lee\textsuperscript{1,2\textdagger}%
\thanks{\textsuperscript{*}Equal contribution.}%
\thanks{\textsuperscript{\textdagger}Corresponding author: Kimin Lee {\tt\small kiminlee@kaist.ac.kr}}%
\thanks{\textsuperscript{1}Kim Jaechul Graduate School of AI, KAIST, Seoul, Republic of Korea.}%
\thanks{\textsuperscript{2}Config, Seoul, Republic of Korea.}%
}
\maketitle

\begin{abstract}
For natural human-robot interaction, a robot must understand human intent expressed not only through language but also through nonverbal signals such as gestures and gaze. However, robot policies still take language instructions as the sole interface for conveying intent, leaving nonverbal signals unused and placing the full burden of communication. In this work, we present \metabbr, a robot framework that captures the human's nonverbal signals through continuous streams of first-person view and gaze from smart glasses, and uses them alongside language instructions as inputs to the robot policy. Our hardware system streams the human's first-person view, gaze, and speech to the robot in real time, transcribing the speech into language instructions. To handle these rich but noisy signals, we design a hierarchical policy in which a high-level policy infers the human's intent and produces a sequence of subtasks, where each subtask is represented as a fine-grained instruction paired with a keyframe that grounds the intent in the scene (e.g., the frame where the human points at the target object). A low-level policy then executes these subtasks. In our experiments on human-robot interactive tasks, \metabbr enables the robot to act on the human's nonverbal signals even when intent is expressed only briefly. Visit our \href{https://project-edith.github.io}{\texttt{project page}} for source code and real-robot demo videos.
\end{abstract}

\section{Introduction}

Language offers humans an intuitive and flexible interface for conveying intent to robots, motivating a growing body of work on language-conditioned policies for robot control~\cite{mees2022calvin, lynch2020language, stepputtis2020language}. Recent advances in multi-modal encoders~\cite{radford2021learning, tschannen2025siglip}, Vision-Language Models (VLMs)\cite{liu2023visual, beyer2024paligemma}, and text-to-video generation models~\cite{cosmospredict, wan2025wan} pretrained on internet-scale data have substantially expanded the generalization of such policies, equipping them with the rich visual-linguistic priors needed to follow free-form instructions~\cite{bc-z, shridhar2022cliport, octo, lbm, pi0.7, ye2026world}.

\begin{figure}[t]
    %\vspace{-1.2\baselineskip}
    \centering
    \includegraphics[
        width=0.4\textwidth
    ]{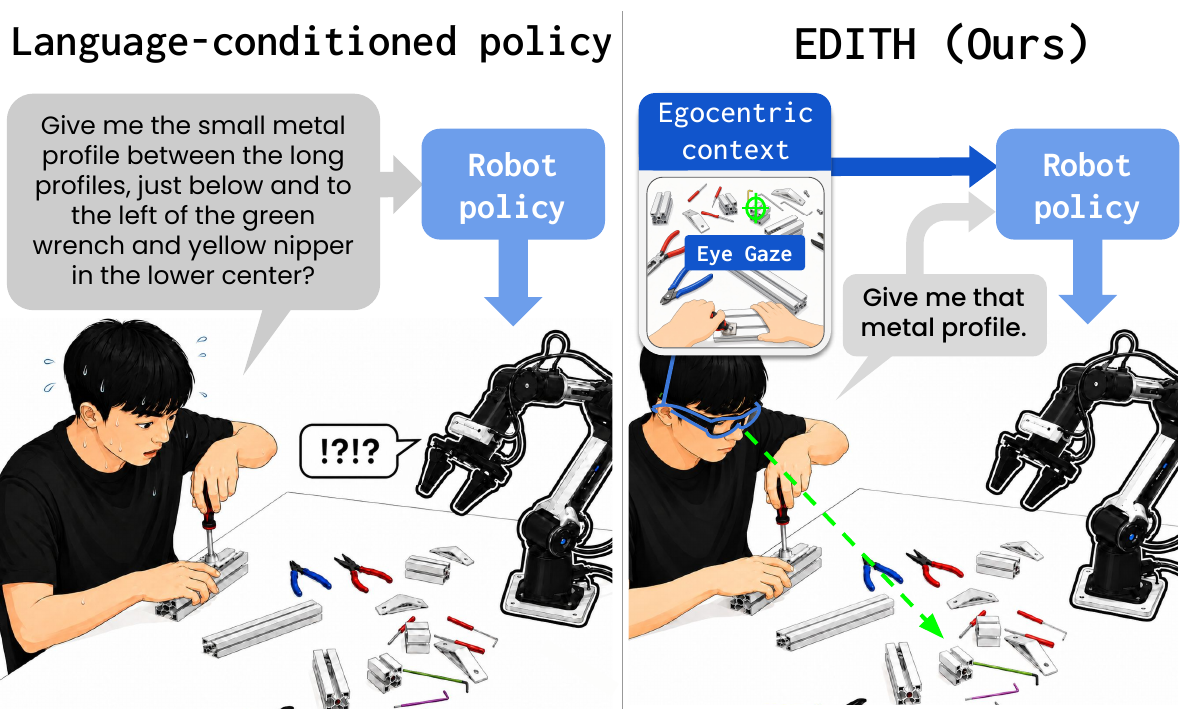}
    \caption{\textbf{(Left)} Language-conditioned policies require fully verbalized text, which is often cumbersome and imprecise. \textbf{(Right)} We leverage human egocentric view and gaze to capture nonverbal signals for natural human-robot interaction.}
    \label{fig:overview}
    \vspace{-0.6cm}
\end{figure}

However, language alone captures only part of how humans naturally express intent. In everyday interaction, people rarely produce fully specified verbal instructions; instead, they ground their speech in nonverbal cues such as eye gaze and gestures~\cite{holler2019multimodal, matuszek2014learning}. As illustrated in Figure~\ref{fig:overview}, a person asking for one of several metal profiles scattered across a table will typically pair a brief verbal request with a glance or pointing gesture toward the desired piece, rather than describing it in detail.
To support such interactions, a robot must jointly interpret a human's verbal and nonverbal signals as they arrive, infer the underlying intent, and act on it.

In this work, we present \metabbr, a robot framework that enables natural human-robot interaction by incorporating live streams of the human's first-person view and eye gaze, in addition to verbal instructions as inputs to the control policy (Figure~\ref{fig:overview}). To capture these signals, our hardware system is based on Project Aria glasses~\cite{engel2023project}, which stream the wearer's egocentric view images, gaze coordinates, and speech. On top of this hardware, we design a hierarchical policy that pairs a VLM as the high-level controller with a Vision-Language-Action (VLA) model as the low-level controller. The high-level policy infers intent from the egocentric streams together with the user's speech, and produces subtasks, each represented as a fine-grained instruction (e.g., \emph{``Pick up the driver with red handle and pass it to the human''}) paired with a keyframe retrieved from the stream that visually grounds the relevant nonverbal cue (e.g., the frame in which the human is gazing at the target driver). The low-level policy then conditions on this instruction-keyframe pair to produce actions.

% evaluate on 3 tasks
% Baselines depends solely on langauge interface fails, 
% Even modular baseline that leverages human's eye gaze stream (FAM-HRI) fails, but 
We evaluate \metabbr on three challenging human-robot interaction tasks, including assisting humans with assembly and serving muffins, in which the human conveys intent through both verbal signals (coarse language instructions) and nonverbal signals (e.g., pointing or gazing at objects). 
We find that baselines relying solely on language, including $\pi_{0.5}$~\cite{pi0.5} and a hierarchical policy with a Gemini-3.1-Flash-Lite~\cite{gemini3.1flashlite} planner, fail to capture human intent (below 7\% success rate on average), whereas \metabbr achieves 48.6\% by grounding its actions in the user's intent. 
\metabbr further remains robust when the human's attention drifts to things unrelated to the robot's task in the middle of the interaction.
We believe our work demonstrates the promise of leveraging nonverbal signals together with verbal instructions to enable natural human-robot interaction.

\section{Related Work}
\label{sec:related_work}

\textbf{Language-based interaction for robot control.} Language has become a key interface for conveying user intent to robots~\cite{tellex2020robots,mavridis2015review,hunt2024survey}.
Early systems mapped structured commands to predefined actions~\cite{liu2017systems,bugmann2004corpus}, but struggled with varied and ambiguous language~\cite{deits2013clarifying,thomason2015learning,thomason2020jointly}. 
Internet-scale pretrained models, equipped with general world knowledge, have enabled new modes of language-based control. LLMs generate executable robot code from natural-language instructions~\cite{liang2023code,singh2022progprompt},
while VLA models pair these with robot data to map instructions directly to actions~\cite{pi0.5, pi0, rt1, openvla, rt2, gr00t-n1}.
To handle more complex instructions, recent works adopt hierarchical architectures in which a VLM serves as a high-level planner and a VLA as a low-level executor~\cite{shi2025hi}.

\textbf{Leveraging nonverbal signals for robot control.} To enable natural human-robot interaction, prior works leverage nonverbal cues such as deictic gestures, gaze, hand pose, and body motion to infer the user's intended target and guide robot actions~\cite{matuszek2014learning, lin2023giraf, admoni2016predicting, su2023recent}.
More recently, smart glasses have been used to capture the user's gaze and egocentric view, which are processed by VLMs to generate robot control code~\cite{fam-hri, tay2026intent}.
However, built as modular pipelines of gaze fixation detectors, hand keypoint extractors, and object detectors, these systems often struggle with weak or transient nonverbal signals such as brief gaze shifts or quickly vanishing gestures~\cite{lin2023giraf, tay2026intent}. Moreover, these systems map the inferred intent to predefined action primitives rather than to a learned policy, which limits them to the motions the primitives can express.
In contrast, \metabbr passes the inferred intent to a VLA policy trained on our demonstrations, so the robot can perform the closed-loop manipulation that natural interaction requires, such as placing an object on a human-held plate.
\section{Background}
\label{sec:background}

\begin{figure*}[t]
    \centering
    \includegraphics[width=0.8\textwidth]{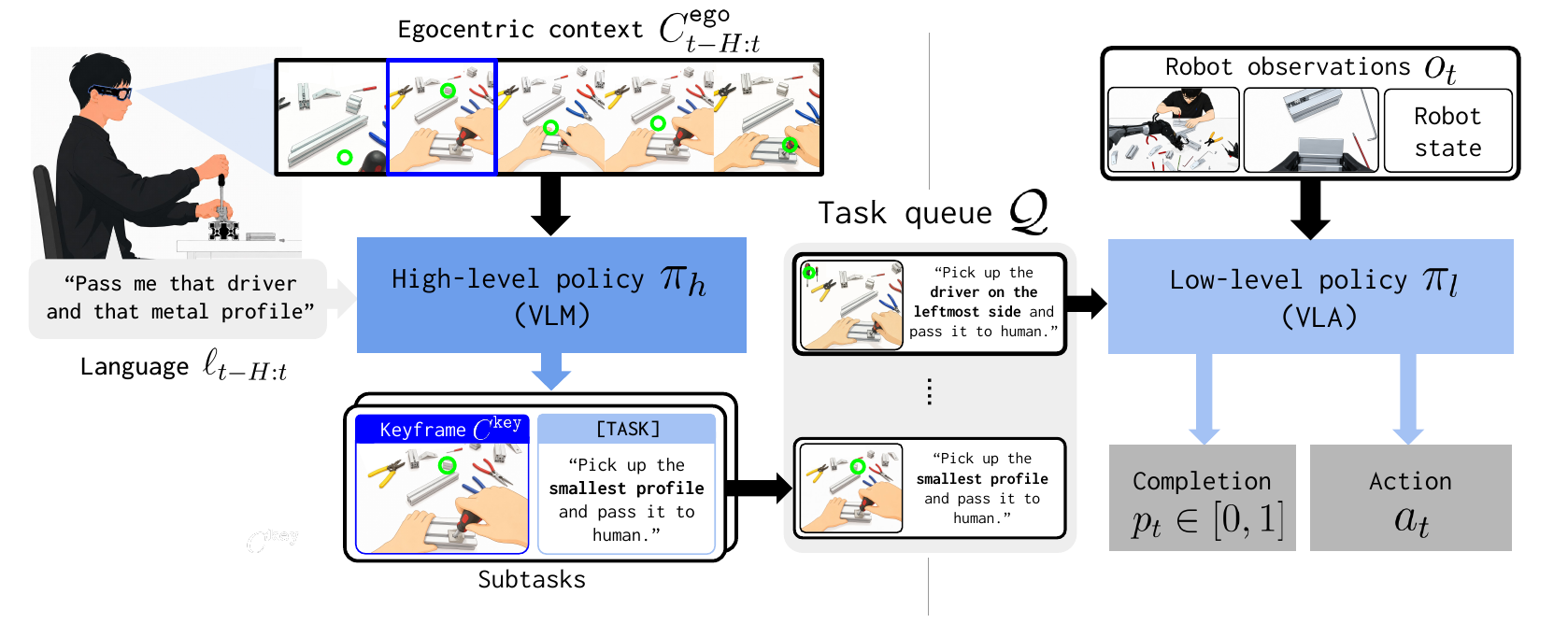}
    \vspace{-0.2cm}
    \caption{Overview of \metabbr. The high-level policy $\pi_h$ processes the egocentric context $C^\texttt{ego}_{t-H:t}$ and language instructions $\ell_{t-H:t}$ to produce subtasks, where each subtask is a pair of a fine-grained instruction $\texttt{[TASK]}$ and a keyframe $C^\texttt{key}$ from $C^\texttt{ego}_{t-H:t}$. These subtasks are queued in $\mathcal{Q}$ and the low-level policy $\pi_l$ executes them by producing robot actions $a_t$.}
    % \caption{%\dk{The high-level policy processes multi-modal human requests to infer the human's intent and generate subtask directives. The low-level policy produces robot actions conditioned on the synchronized robot observations abd generated subtask directives,}
    % Overview of the inference procedure. The high-level policy $\pi_h$ processes the egocentric context $C^\texttt{ego}_{t-H:t}$ and language $\ell_{t-H:t}$ to produce subtasks, each pairing an subtask instruction $\texttt{[TASK]}$ with a keyframe $C^\texttt{key}$ retrieved from $C^\texttt{ego}_{t-H:t}$. The subtasks are pushed into a task queue $\mathcal{Q}$, from which the low-level policy $\pi_l$ pops and executes them as robot actions $a_t$.
    % %\dk{A high-level policy $\pi_h$ periodically infers human's intent from a  recent window of egocentric contexts and produces subtasks, ..
    % %The subtasks are pushed into task queue $\mathcal{Q}$
    % %human requests to infer the human's intent and generate subtask directives. The low-level policy produces robot actions conditioned on the synchronized robot observations abd generated subtask directives,}
    % }
    \label{fig:policy_design}
    \vspace{-0.1in}
\end{figure*}
%We first describe the formulation of goal-conditioned policy for robot control and then discuss the problem that motivates our framework.

%\paragraph{Formulation of robot control.}
%Robot control can be formulated as a Partially Observable Markov Decision Process (POMDP), defined by $<\mathcal{S}, \mathcal{O}, \mathcal{A}, \mathcal{P}, \mathcal{R}>$.
\textbf{Language-conditioned robot policy.}
We formulate robot control as a sequential decision-making problem. At each discrete timestep $t$, given an observation $o_t$ (e.g., RGB images from cameras and the robot's proprioceptive state) and a language instruction $\ell_t$, a language-conditioned policy $\pi(a_t \mid o_t, \ell_t)$ predicts an action $a_t$, executes it on the robot, and receives the next observation $o_{t+1}$ in a closed loop. Here, the language serves as an  interface for conveying human intent, but it captures only an explicit, verbalized slice of communication. We instead consider settings where humans express intent through language, together with nonverbal signals such as eye gaze or gestures.

\textbf{Hierarchical architecture.}
For complex tasks requiring long-horizon robot control, the language-conditioned policy $\pi(a_t \mid o_t, \ell_t)$ can be implemented as a hierarchical architecture composed of a high-level policy $\pi_h$ and a low-level policy.
Analogous to the System~1 / System~2 view of human cognition~\cite{kahneman2011thinking}, $\pi_h$ decomposes a complex task $\ell_t$ into an atomic subtask $g_t$, while $\pi_l$ produces the action $a_t$ conditioned on $g_t$ and the robot observation $o_t$: $\pi(a_t \mid o_t, \ell_t) = \pi_l(a_t \mid o_t, g_t) \, \pi_h(g_t \mid o_t, \ell_t).$
% \begin{equation}
% \pi(a_t \mid o_t, \ell_t) = \pi_l(a_t \mid o_t, g_t) \, \pi_h(g_t \mid o_t, \ell_t).
% \label{eq:hierarchical_policy}
% \end{equation}
Large Language Models (LLMs) and VLMs are commonly used as $\pi_h$ for their reasoning capabilities, while Vision-Language-Action (VLA) models~\cite{pi0.5, pi0} serve as $\pi_l$.
Prior work represents $g_t$ as either a fine-grained instruction~\cite{pi0.5, shi2025hi, ahn2022can} or a subgoal image~\cite{pi0.7, hamster, susie}.

\section{\metabbr}

In this section, we introduce \metabbr, a robot framework that leverages both the human's verbal (i.e., language instruction) and nonverbal signals (e.g., gestures, gaze, ongoing activities) for more natural human-robot interaction (see Figure~\ref{fig:overview}). 
%\dj{We represent the nonverbal context through the human's first-person view and gaze.}
\metabbr consists of a smart-glasses-based hardware system that captures the human's first-person view, gaze, and speech in real time (Section~\ref{subsec:hardware_design}), and a hierarchical policy that processes these signals to produce robot actions (Section~\ref{subsec:policy_architecture}). Finally, we describe data collection and policy training scheme in Section~\ref{subsec:data_collection_and_training}.

\subsection{Hardware System}
\label{subsec:hardware_design}

We build a hardware system that streams the human's first-person view, gaze, and speech in real time, and synchronizes them with the robot's observations.

\textbf{Why first-person view and gaze?}
We first motivate our choice of first-person view and gaze as the representation of the human's nonverbal signals. The first-person view visually captures the human's gestures and ongoing activities, while gaze reveals where their attention is focused within the visual frame. Together, these two streams provide cues for a robot policy to infer the human's underlying intent and needs. We hereafter denote the first-person view image and gaze coordinates at timestep $t$ as the egocentric context $C_t^{\texttt{ego}}$.

\textbf{Capturing human signals via smart glasses.} To capture the human's egocentric context $C_t^{\texttt{ego}}$ and language instruction $\ell_t$ in real time, we use Project Aria glasses~\cite{engel2023project}, a lightweight wearable device with a built-in first-person RGB camera, eye tracker, and microphone. The glasses stream the first-person view, gaze, and speech to the robot server, which transcribes speech into the language instruction $\ell_t$ and synchronizes all signals with the robot observation $o_t$ by timestamp, producing an aligned stream of $(C_t^{\texttt{ego}}, \ell_t, o_t)$ at each timestep $t$. %Further details are provided in Appendix~\ref{app:impl_detail_hardware}.

\subsection{Policy Design}
\label{subsec:policy_architecture}
Given $(C_t^{\texttt{ego}}, \ell_t, o_t)$, we now describe how \metabbr uses these signals to produce actions. 
While the human signals $(C_t^{\texttt{ego}}, \ell_t)$ are rich, they are also noisy: humans often express their intent briefly, interleaved with unrelated ongoing activities.
We empirically find that directly feeding these signals to a single end-to-end policy fails to produce actions that follow the human's underlying intent (see Section~\ref{subsec:main_result} for supporting results).
We therefore design a hierarchical policy that decouples the problem of inferring human intent from producing low-level actions.

\textbf{Overview of the hierarchical policy.} The hierarchical policy consists of a high-level policy $\pi_h$ and a low-level policy $\pi_l$.
$\pi_h$ periodically infers intent from the human's egocentric context stream and produces a sequence of subtasks for $\pi_l$ to execute.
Each subtask is represented as a pair of a fine-grained instruction and a keyframe retrieved from the input stream, where the keyframe captures the moment the human's nonverbal signal is most clearly expressed.
For example (Figure~\ref{fig:policy_design}), when the human says ``Pass me that driver and that metal profile'' while pointing at each, $\pi_h$ produces two subtasks, each pairing an instruction (e.g., ``Pick up the driver with a hexagonal tip and pass it to the human'') with the keyframe where the human's gaze fixates on the target.
The keyframe grounds nonverbal signals that language alone often fails to specify, giving $\pi_l$ a precise and compact subtask specification.
Conditioned on both, $\pi_l$ produces low-level robot actions and a completion probability; once this probability exceeds a threshold, $\pi_l$ proceeds to the next subtask.

While the above describes how a single subtask is represented and executed, real-time human-robot interaction poses two additional challenges.
First, the human may express intent at any moment, even while $\pi_l$ is executing a previous subtask, so $\pi_h$ must monitor human signals in parallel with $\pi_l$.
Second, the human may express multiple intents at once (e.g., ``Give me these things'' while pointing at several targets), producing multiple subtasks that must be queued for sequential execution.
To meet both requirements, we run $\pi_h$ and $\pi_l$ asynchronously, connecting them through a task queue $\mathcal{Q}$: $\pi_h$ appends newly identified subtasks to $\mathcal{Q}$, while $\pi_l$ pops and executes the subtask at its head.
We describe the full procedure in Algorithm~\ref{alg:inference}.

\textbf{High-level policy.} $\pi_h$ infers intent from the streams of human signals $(C_t^{\texttt{ego}}, \ell_t)$. Specifically, every $H$ timesteps, $\pi_h$ takes as input the recent $H$ frames of egocentric context $C_{t-H:t}^{\texttt{ego}}$ together with the language instructions $\ell_{t-H:t}$ captured within this window, and produces a set of subtasks:
\begin{equation}
\{(\texttt{[TASK]}_i, C_i^{\tt key})\}_{i=1}^n \sim \pi_h(\cdot \mid C_{t-H:t}^{\texttt{ego}}, \ell_{t-H:t}),
\label{eq:high_level_policy}
\end{equation}
where each $\texttt{[TASK]}_i$ is a fine-grained language instruction specifying the $i$-th subtask, and $C_i^{\texttt{key}}$ is the keyframe in $C_{t-H:t}^{\texttt{ego}}$ at which the human's nonverbal signal for $\texttt{[TASK]}_i$ is most clearly expressed. These subtasks are then pushed into $\mathcal{Q}$. When no intent is expressed through either language or nonverbal signals within the window, $\pi_h$ returns the empty set ($n=0$), leaving $\mathcal{Q}$ unchanged.

We implement $\pi_h$ with a VLM (e.g., Gemini-3.1-flash-lite~\cite{gemini3.1flashlite}), leveraging its world knowledge to infer human intent from multimodal context~\cite{choi2026state}. To feed $C_{t-H:t}^{\texttt{ego}}$ and $\ell_{t-H:t}$ jointly to the VLM, we render them as a single video clip by overlaying gaze markers on each first-person RGB frame and inserting transcribed speech as time-aligned captions.\footnote{We obtain word-level timestamps for the captions using the Whisper API~\cite{radford2023robust}.} For each subtask, the VLM outputs a fine-grained instruction $\texttt{[TASK]}$ and a timestamp index, and we retrieve the frame at this index from $C_{t-H:t}^{\texttt{ego}}$ as $C^{\texttt{key}}$. %Further details are in Appendix~\ref{app:impl_detail_policy}.

%For each subtask, rather than directly generating the keyframe, the VLM outputs a fine-grained language instruction $\texttt{[TASK]}$ along with a timestamp index, and we retrieve the frame at this index from $C_{t-H:t}^{\texttt{ego}}$ as $C^{\texttt{key}}$. This grounds each subtask in the actual observed scene rather than a synthesized image. Further implementation details are provided in Appendix~\ref{app:impl_detail_policy}.

\textbf{Low-level policy.} The low-level policy $\pi_l$ sequentially executes each subtask in $\mathcal{Q}$ at the robot control frequency. Given the subtask $(\texttt{[TASK]}, C^{\texttt{key}})$ at the head of $\mathcal{Q}$ and the robot observation $o_t$, $\pi_l$ produces a low-level robot action $a_t$ together with a completion probability $p_t \in [0, 1]$:
\begin{equation}
(a_t, p_t) \sim \pi_l(\cdot \mid o_t, \texttt{[TASK]}, C^{\texttt{key}}).
\label{eq:low_level_policy}
\end{equation}
Here, $p_t$ is $\pi_l$'s estimate of whether the current subtask has been completed; once $p_t$ exceeds a threshold, $\pi_l$ pops the subtask from $\mathcal{Q}$ and proceeds to the next. If $\mathcal{Q}$ is empty, $\pi_l$ remains idle until $\pi_h$ pushes a new subtask.
We implement $\pi_l$ by fine-tuning a pre-trained VLA model (e.g., $\pi_{0.5}$~\cite{pi0.5}) on our dataset (Section~\ref{subsec:data_collection_and_training}). To predict $p_t$, we add a linear head on top of the VLA backbone.

\begin{algorithm}[t]
\small
\caption{Inference of the \metabbr Policy}
\label{alg:inference}
\KwIn{High-level policy $\pi_h$, low-level policy $\pi_l$, context window size $H$, max horizon $T$, threshold $\tau$}
\BlankLine
Initialize shared task queue $\mathcal{Q} \gets \emptyset$ \;
\textbf{Run \textcolor{softblue}{Process 1} and \textcolor{softblue}{Process 2} asynchronously.}
\BlankLine
\textbf{\textcolor{softblue}{Process 1}: $\pi_h$ monitors the human's egocentric context every $H$ steps and produces subtasks.} \\
$t \gets 1$ \;
\While{$t < T$}{
    \If{$t \bmod H = 0$}{
        Recent $H$ frames of egocentric context as $C_{t-H:t}^\texttt{ego}$ \;
        User's instruction in recent $H$ frames as $\ell_{t-H:t}$ \;
        $\{(\texttt{[TASK]}_i, C_i^\texttt{key})\}_{i=1}^n \gets \pi_h(C_{t-H:t}^{\texttt{ego}}, \ell_{t-H:t})$\\
        $\mathcal{Q}.\texttt{append}(\{(\texttt{[TASK]}_i, C_i^\texttt{key})\}_{i=1}^n)$ \\
    }
    $t \gets t + 1$ \;
}
\BlankLine
\textbf{\textcolor{softblue}{Process 2}: $\pi_l$ executes subtasks queued in $\mathcal{Q}$} \\
$t \gets 1$ \;
Reset the robot and read its initial observation $o_t$ \;
\While{$t < T$}{
    \If{$\mathcal{Q} = \emptyset$}{
        \textbf{await} until $\mathcal{Q} \neq \emptyset$ \\
    }
    $\texttt{[TASK]}, C^\texttt{key} \gets \texttt{head}(\mathcal{Q})$\\
    $(a_t, p_t) \gets \pi_l(o_t, \texttt{[TASK]}, C^\texttt{key})$ \\
    Execute $a_t$ on the robot and observe $o_{t+1}$ \;
    \If{$p_t > \tau$}{
        $\text{Pop}(\texttt{[TASK]}, C^\texttt{key})$ from $\mathcal{Q}$ \\
    }
    $t \gets t + 1$ \;
}
\end{algorithm}

\subsection{Data Collection and Policy Training}
\label{subsec:data_collection_and_training}

\textbf{Data collection.} To train our hierarchical policy, we collect a demonstration dataset in human-robot interaction scenarios, where a human conveys intent to a robot through language instructions and nonverbal signals such as gestures and gaze (see Figure~\ref{fig:evaluation_tasks} for example tasks). Each episode involves two participants: a human actor and a robot teleoperator. The human actor, wearing Project Aria glasses, expresses intent to the robot through speech, gestures, and gaze (e.g., gazing at a specific driver and saying ``give me that driver'' while assembling a metal stand). The teleoperator then controls the robot to fulfill the intent (e.g., picking up the requested driver and passing it to the human). Throughout the interaction, the human's egocentric context $C_t^{\texttt{ego}}$, language instruction $\ell_t$, robot observation $o_t$, and robot action $a_t$ are synchronously recorded, yielding a trajectory $\tau = \{C_t^{\texttt{ego}}, \ell_t, o_t, a_t\}_{t=1}^L$. %Further details of the data collection protocol are provided in Appendix~\ref{app:data_curation_raw}.

After collection, we annotate each trajectory $\tau$ in three steps. First, since $\tau$ spans the entire interaction and typically contains multiple consecutive subtasks, we temporally segment $\tau$ into subtasks. %(see Appendix~\ref{app:data_curation_label} for details).
Second, for each segment, we annotate the fine-grained instruction $\texttt{[TASK]}$ and the keyframe $C^{\texttt{key}}$. Finally, within each segment, we label every frame with a completion probability $p_t$ ($p_t = 1$ near the end of the segment and $p_t = 0$ otherwise). This yields a labeled subtask trajectory $\tau_{\texttt{labeled}} = \bigl((\texttt{[TASK]}, C^{\texttt{key}}), \{(o_t, a_t, p_t)\}_{t=1}^M\bigr)$.

\begin{figure}[h]
    \centering     \includegraphics[width=0.3\textwidth]{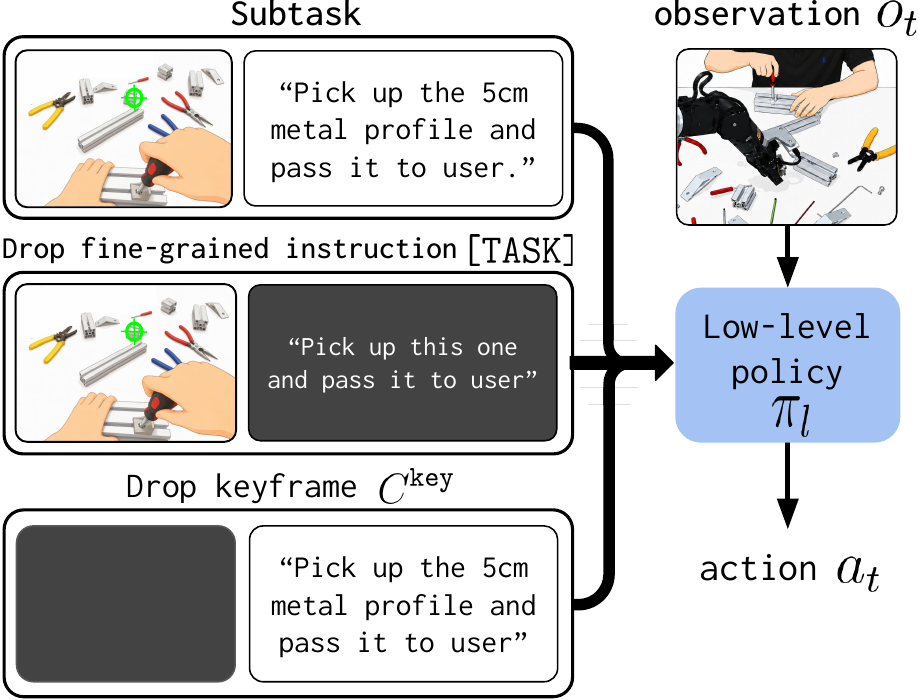}
\caption{\emph{Modality dropout}. We randomly either replace the instruction (middle) into underspecified one or drop the keyframe (bottom), enforcing $\pi_l$ to use both modalities.}
    %\caption{Example of the modality dropout. During training, we randomly drop either subtask instruction or keyframe, in order to encourage $\pi_l$ learns to leverage both.}
    \label{fig:modality_dropout}
\end{figure}

\textbf{Policy training.}
While both $\pi_h$ and $\pi_l$ could be trained on $\tau_{\texttt{labeled}}$, we find that closed VLMs (e.g., Gemini-3.1-Flash-Lite) outperform open-source alternatives as $\pi_h$, so we focus on training $\pi_l$ to predict $a_t$ given the robot observation $o_t$ and the subtask $(\texttt{[TASK]}, C^{\texttt{key}})$. We instantiate $\pi_l$ as a VLA model with a flow-matching action head~\cite{pi0.5}, and train it with the flow-matching objective~\cite{pi0} by feeding $C^{\texttt{key}}$ as a first-person image with the gaze marker overlaid. We also train a linear head on the frozen VLA backbone to predict $p_t$ with a binary cross-entropy loss.

Naively training $\pi_l$ on both modalities biases it toward relying on only one, since either is often sufficient for action prediction. To force $\pi_l$ to use both, we introduce \textit{modality dropout} (Figure~\ref{fig:modality_dropout}). For each training sample, we choose one of three options with equal probability: (i) keep both modalities intact; (ii) replace $\texttt{[TASK]}$ with an underspecified version (e.g., ``pick up this one'' instead of ``pick up the 5cm metal profile''), forcing reliance on $C^{\texttt{key}}$; or (iii) replace $C^{\texttt{key}}$ with a blank image, forcing reliance on $\texttt{[TASK]}$. 

\section{Experiments}
\label{sec:experiment}

Our experiments answer three questions: (1) Does \metabbr enable the robot to interpret and respond to the human's nonverbal signals? (Section~\ref{subsec:main_result}) (2) How does each core component of \metabbr contribute, and does it remain robust to distractions during interaction? (Section~\ref{subsec:ablation}) (3) What are the main failure modes of \metabbr? (Section~\ref{subsec:failure})

\subsection{Setup}
\label{subsec:setup}

\begin{figure}[h]
    \centering
    \includegraphics[width=0.47\textwidth]{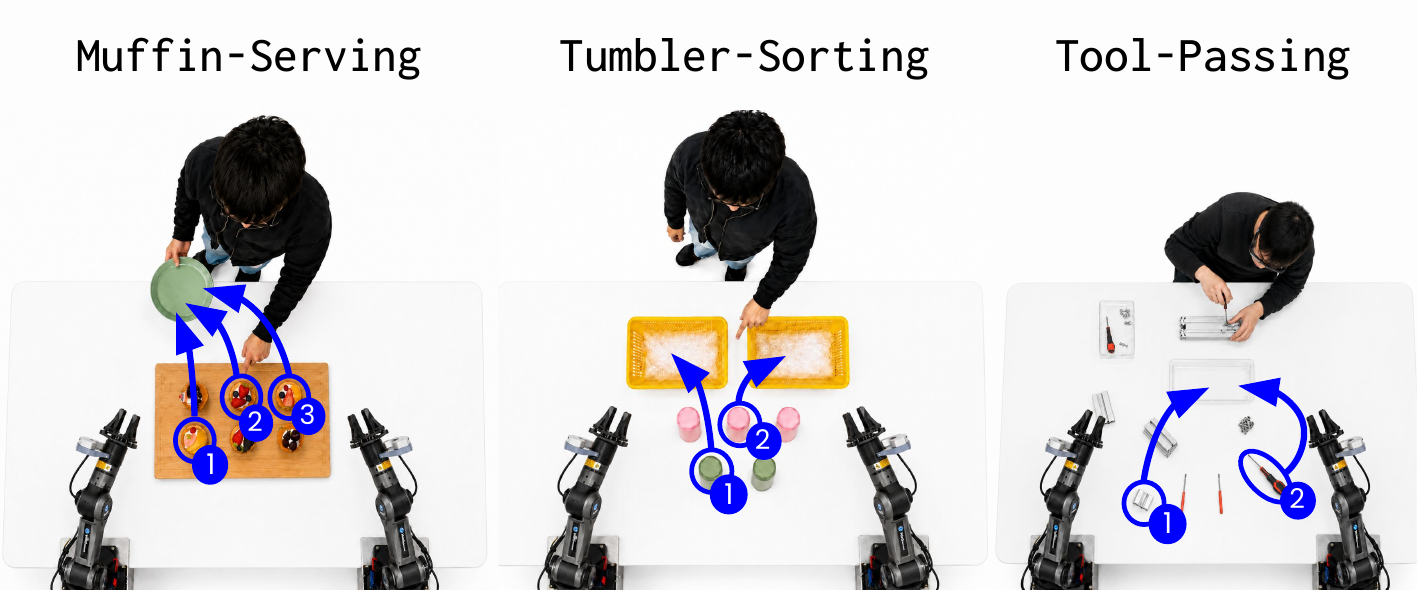}
    \caption{Three human-robot interaction tasks. Numbered blue arrows indicate the target object and order of each request.} %없애면 한줄 생김
    \label{fig:evaluation_tasks}
\end{figure}
\textbf{Tasks.} We design three evaluation tasks involving human-robot interaction, where a human evaluator expresses intent in real time through language instruction complemented by nonverbal signals.
In every task, the language instruction is insufficient on its own, requiring the robot to interpret the human's nonverbal signals to succeed.
Figure~\ref{fig:evaluation_tasks} illustrates the tasks, which we detail below:
\begin{itemize}
\setlength{\leftskip}{-0.35cm}
    \item \texttt{Muffin-Serving} : 
    Six muffins are densely arranged in a display, and the human requests three of them through a single instruction (e.g., ``Give me this muffin, this muffin, and this muffin'') while pointing at each one.
    The robot must understand the gesture and serve the corresponding muffins to a plate held by the human.
    \item \texttt{Tumbler-Sorting} : Five tumblers and two baskets are placed on the table. The human requests placing two tumblers into different baskets through a single instruction (e.g., ``Put this tumbler to that basket, this tumbler to that basket'') while pointing at each tumbler and basket. The robot must ground these gestures and place each tumbler into the corresponding basket.
    \item \texttt{Tool-Passing} : While assembling a stand, the human requests tools (drivers and metal profiles) as they are needed, conveying each request through a verbal instruction paired with gaze (e.g., glancing at a specific driver while saying ``Give me that driver''). The human first requests a particular driver from among many on the table, uses it to continue the assembly, and later requests a specific metal profile in the same manner.
    The robot must respond to these intermittent requests and ground the intended target from the human's gaze.
\end{itemize}

\begin{figure*}[ht]
    \centering
    \includegraphics[width=0.95\textwidth]{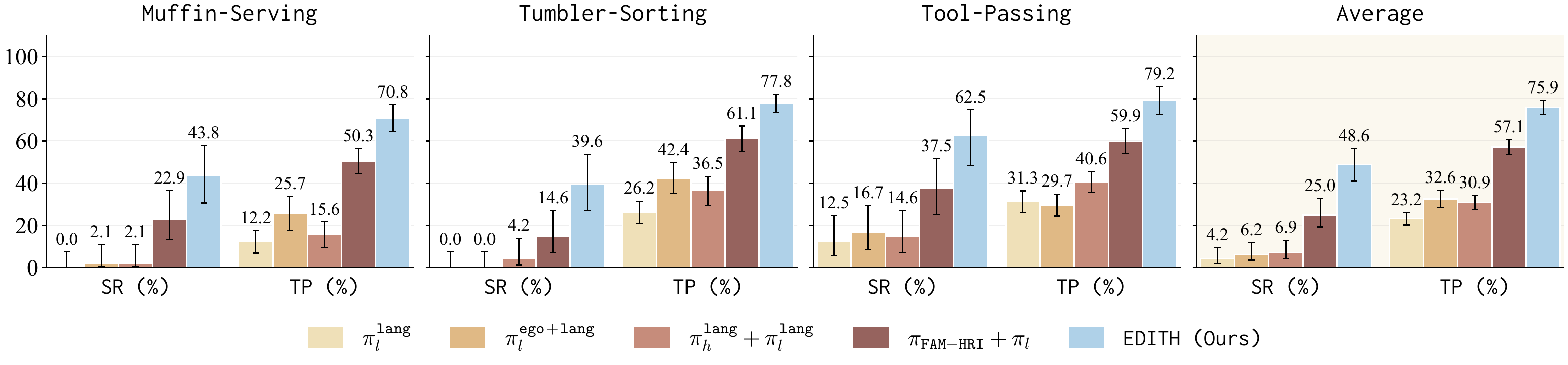}
    \caption{Success rate (SR) and task progress (TP) of \metabbr compared against baselines. Error bars indicate the 95\% confidence interval of the reported SR / TP estimates, computed over 48 trials evaluated by 4 humans. In each trial, the human conveys intent through an underspecified language instruction containing referring expressions (e.g., ``Give me this muffin, this muffin, and this muffin'') accompanied by nonverbal cues such as pointing gestures.} %Error bars show the standard error of a binomial proportion for SR and the standard deviation for TP.}
    \vspace{-0.5cm}
    \label{fig:main_result}
\end{figure*}

\textbf{Evaluation protocol.}
We evaluate \metabbr and baselines using two metrics: success rate (SR) and task progress (TP).
SR is the fraction of trials in which all subtasks are completed, whereas TP is the average fraction of subtasks completed per trial.
For evaluation, we recruit four human evaluators: two who participated in collecting the training data and two who did not.
Each evaluator runs 12 trials per task per method, where each trial uses a configuration unseen during training, differing in the arrangement of the objects and in which objects the human requests.
We then report SR and TP averaged over the four evaluators.
%We run 48 trials per task per method and report the mean SR and TP.

\textbf{Implementation details.}
We collect 120 demonstrations for \texttt{Tool-Passing} and 80 for each of the other tasks (Section~\ref{subsec:data_collection_and_training}), and finetune $\pi_{0.5}$ on each task for 5 epochs with AdamW (learning rate $10^{-5}$, batch size 32).
At inference, $\pi_h$ uses a context window of $H=120$, and $\pi_l$ uses a completion threshold of $\tau=0.8$.
We use a bimanual 6-DoF Agilex Piper with parallel-jaw grippers and three RealSense RGB cameras.

\textbf{Tested methods.}
For our method, we use Gemini-3.1-Flash-Lite~\cite{gemini3.1flashlite} as the high-level policy $\pi_h$, and finetune a pretrained $\pi_{0.5}$~\cite{pi0.5} on our subtask-segmented demonstrations as the low-level policy.
We compare \metabbr against four baselines: %(see Appendix~\ref{appsub:baselines} for details): 

\textbf{(i)} $\pi_l^\texttt{lang}$ is an end-to-end language-conditioned policy that takes as input the robot observation and a language instruction. We implement this by finetuning $\pi_{0.5}$ on the same demonstrations without subtask segmentation, conditioned on a language instruction that fully specifies the task (e.g., ``Give me a strawberry muffin, a cherry muffin, and an Oreo muffin.''). 

\textbf{(ii)} $\pi_h^\texttt{lang}\text{+}\pi_l^\texttt{lang}$ is a hierarchical policy akin to Hi-Robot~\cite{shi2025hi}, where Gemini-3.1-Flash-Lite acts as the high-level policy, taking the language instruction and robot-camera RGB streams as input to produce a subtask instruction for the low-level policy. The low-level policy is a $\pi_{0.5}$ finetuned on the same subtask-segmented demonstrations as our method, but conditioned on the subtask instruction without the keyframe. This baseline differs from \metabbr only in that it omits the human's egocentric context and the subtask keyframe.

\textbf{(iii)} $\pi_l^\texttt{ego+lang}$ is an end-to-end policy that takes as input the human's current egocentric context $C_t^\texttt{ego}$ in addition to the robot observation and language instruction. We finetune $\pi_{0.5}$ on the same unsegmented demonstrations as $\pi_l^\texttt{lang}$, additionally conditioned on $C_t^\texttt{ego}$. During training, we apply the modality dropout described in Section~\ref{subsec:data_collection_and_training}; otherwise the policy is biased to ignore $C_t^\texttt{ego}$ and fails when the language instruction is underspecified.

\textbf{(iv)} $\pi_\texttt{FAM-HRI}+\pi_l$ is our adaptation of FAM-HRI~\cite{fam-hri}, a modular pipeline for gaze-based robot manipulation.
Given the human's spoken instruction and streams of eye gaze captured by AR glasses, FAM-HRI first detects the moment of gaze fixation during the utterance as the keyframe. To detect the keyframe, it uses an LLM to identify the intervals in which a referring expression (e.g., ``this'' or ``there'') is spoken, and detects the moment the gaze fixates within each interval. It then locates the object under the gaze in the keyframe via visual feature matching, and executes the action by calling predefined action primitives on the located object.
$\pi_\texttt{FAM-HRI}+\pi_l$ adopts only its keyframe extraction pipeline using Gemini-3.1-flash-lite as an LLM, and pair it with the same low-level policy $\pi_l$ as \metabbr.
%This baseline thus differs from \metabbr only in that the keyframe is selected by FAM-HRI's fixed gaze-fixation rule, rather than by a VLM reasoning over the stream of speech, gaze, and gestures.

\begin{figure}[h]
    \centering
    \includegraphics[width=0.3\textwidth]
    {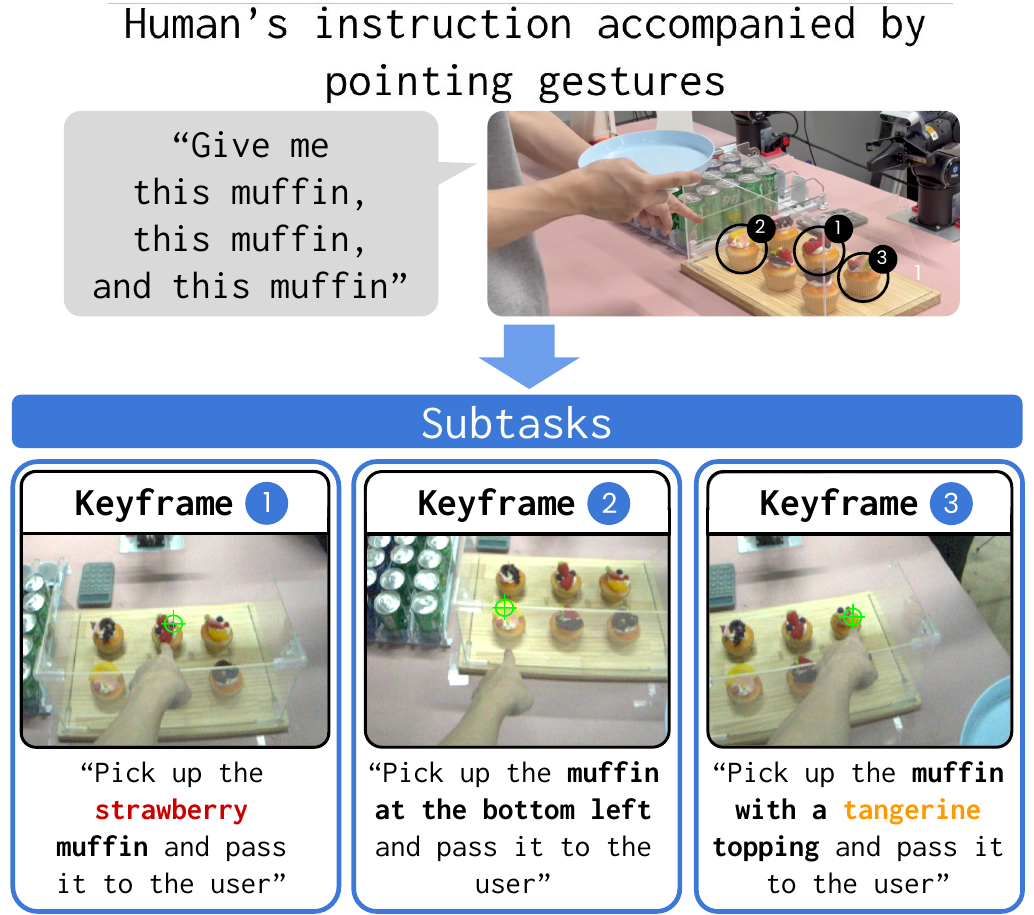}
    \caption{Examples of subtasks produced by the high-level policy of \metabbr on the muffin-serving task.}
    \label{fig:subtask_examples}% 아래쪽 여백 당김 (선택)
\end{figure}

\begin{figure}[h]
    \centering
    \includegraphics[width=0.49\textwidth]{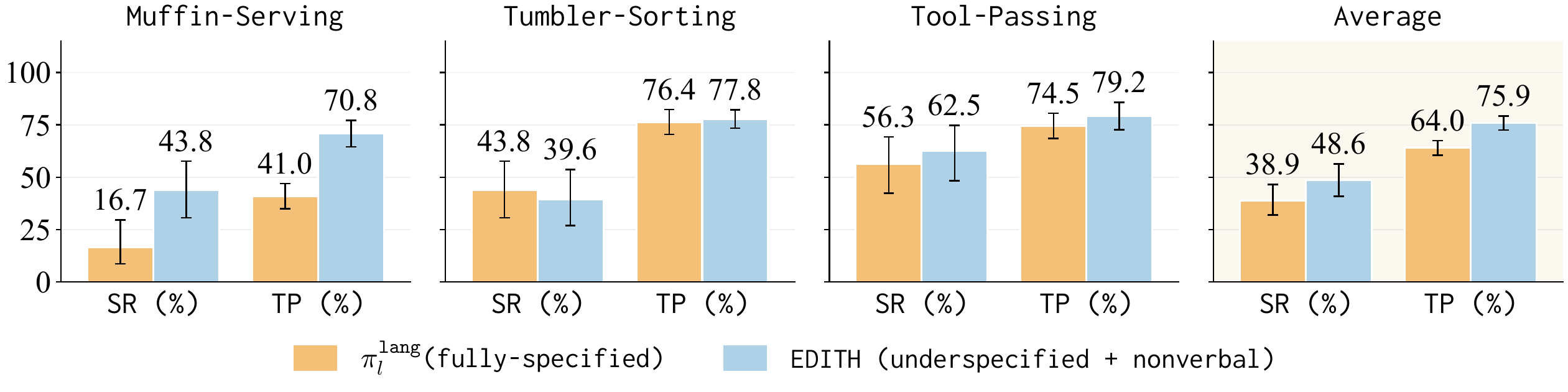}
    \caption{Despite receiving only underspecified instructions that rely on nonverbal cues, \metabbr outperforms $\pi_l^\texttt{lang}$ given fully-specified instructions seen during training, by effectively leveraging egocentric context.}
    \vspace{-0.3cm}
    \label{fig:comparison_to_vla}
\end{figure}

\subsection{Main Results}
\label{subsec:main_result}

Figure~\ref{fig:main_result} presents the success rate and task progress of \metabbr against baselines across the three evaluation tasks, each requiring interpretation of the human's rapidly evolving, complex nonverbal signals.
Despite this difficulty, \metabbr achieves an average success rate of 48.6\% and task progress of 75.9\% by translating the human's nonverbal signals into a sequence of subtasks, each grounded by a keyframe (see Figure~\ref{fig:subtask_examples} for examples produced when the human requests three muffins by pointing at each).
% benefit of egocentric context.

\textbf{Egocentric context is essential.}
$\pi_l^\texttt{lang}$ and $\pi_h^\texttt{lang}\text{+}\pi_l^\texttt{lang}$, baselines that do not use the human's egocentric context, perform poorly,
highlighting the benefits of egocentric context in grounding nonverbal intent.
%By leveraging the egocentric context, \metabbr outperforms these baselines by a large margin.
Notably, it even surpasses $\pi_l^\texttt{lang}$ given the fully-specified language instructions it was trained on, which reaches only a 38.9\% success rate (see Figure~\ref{fig:comparison_to_vla}), showing that the egocentric context conveys what even a fully-specified instruction cannot.\footnote{Our evaluation tasks involve visually similar objects placed together, such as metal profiles that slightly differ only in their size. In such scenes, even fully specifying the target in language is often insufficient for a robot control policy to distinguish it from the others.}

% fundamental limitation of language interface: 
%Notably, when $\pi_l^\texttt{lang}$ is instead given fully-specified instructions that 
%..., still far below \metabbr, which recieves underspecified instructions together with nonverbal expressions.

\textbf{How to process the egocentric context matters.}
Naively exploiting the egocentric context, however, does not guarantee strong performance. 
$\pi_l^\texttt{ego+lang}$, which directly conditions the policy on the egocentric context, yields inconsistent benefits across tasks.
It shows only marginal improvement over the language-only baselines in \texttt{Muffin-Serving} and \texttt{Tumbler-Sorting}, where the human's gaze tends to remain on the target throughout the interaction (e.g., staring at the desired muffin until the robot grasps it), so the real-time egocentric context provides signals it can directly exploit.
In \texttt{Tool-Passing}, however, the gain diminishes, as the human is mostly engaged in their own activity and only intermittently attends to the target.

$\pi_h^\texttt{FAM-HRI}+\pi_l$ outperforms the other baselines by extracting keyframes from the egocentric context based on gaze fixation, yet it still falls behind \metabbr.
This is because the gaze fixation rule often selects the wrong keyframe.
Specifically, in tasks where the human requests multiple objects in a row, such as \texttt{Muffin-Serving}, the human's gaze tends to precede the corresponding verbal referring expression~\cite{griffin2000eyes}.
Consequently, the gaze fixation detected while a referring word is spoken points elsewhere, as the gaze has already moved on to the next target by then, yielding keyframes with the gaze fixated on the wrong object.
Unlike the fixed rule on gaze, \metabbr extracts keyframes with a VLM using its general knowledge, leading to consistently better keyframe extraction and performance across evaluation tasks.
\subsection{Component Ablation and Robustness}
\label{subsec:ablation}
In this section, we provide comprehensive analyses on \metabbr, covering ablations of its core components, and an analysis on the robustness of \metabbr.

\begin{figure}[h]%{r}{0.46\linewidth}
    \centering   
    \includegraphics[width=0.32\textwidth]{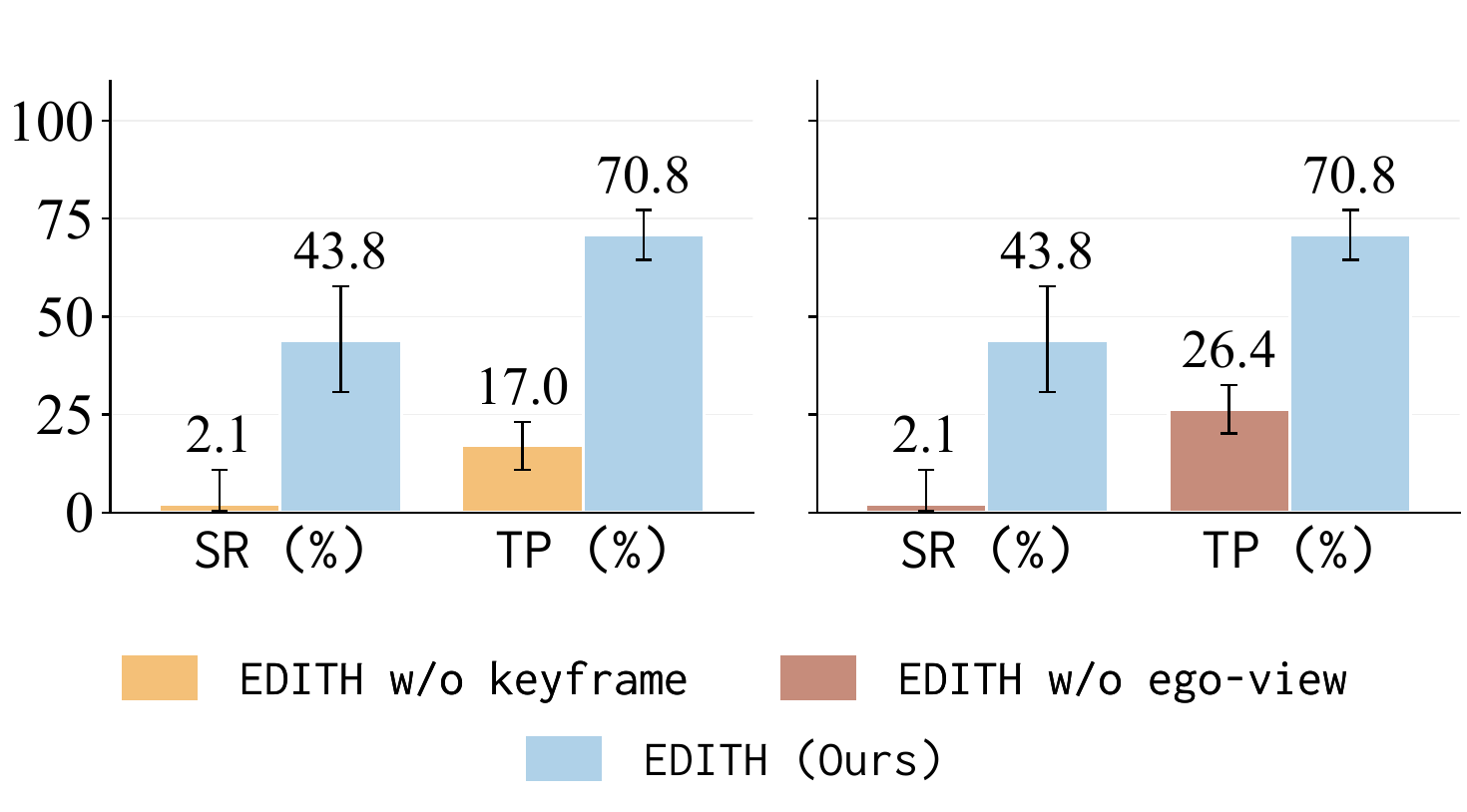}
    \caption{Performance of \metabbr on the \texttt{Muffin-Serving} task, compared against \textbf{(Left)} a baseline without the keyframe and \textbf{(Right)} a baseline that replaces the egocentric view with the robot's view. Removing either component leads to a significant drop in performance.}
    \label{fig:ablation_core_components}                      % 아래쪽 여백 당김 (선택)
    \vspace{-0.3cm}
\end{figure}
\textbf{Ablation of keyframe.}
To isolate the keyframe's contribution, we compare \metabbr against a hierarchical baseline that ablates it in the Muffin-Serving task: the high-level policy $\pi_h$ takes the egocentric context and language instruction to produce subtasks as in \metabbr, while the low-level policy $\pi_l$ is conditioned on the subtask instruction without the keyframe.
As shown in Figure~\ref{fig:ablation_core_components} (Left), removing the keyframe causes a 41.7\%p drop in success rate and 53.8\%p drop in task progress on average, mainly for two reasons. First, $\pi_h$ often hallucinates when translating nonverbal signals into a subtask instruction, producing a description that refers to the wrong target. 
Second, even with a semantically correct instruction, the same subtask can be phrased in multiple equivalent ways, and if $\pi_h$ uses a phrasing unseen during $\pi_l$'s training, $\pi_l$ often fails. Since the keyframe is retrieved from the input egocentric stream rather than generated by $\pi_h$, it avoids both failure modes (i.e., hallucination and ambiguous phrasing).

% \textbf{Ablation of egocentric view.}
% Next, we examine the contribution of leveraging egocentric context. 
% To this end, we compare \metabbr against a hierarchical policy baseline where the high-level policy takes as input the streams of robot-view observation instead of the egocentric context, and produces subtasks each pairing subtask instruction and a keyframe captured from the robot-view stream.
% Different from \metabbr, as a low-level policy, we finetune $\pi_{0.5}$ with the segmented trajectory conditioned on a subtask instruction and a keyframe captured at robot's view.
% In \texttt{Muffin-Serving} task, as shown in Figure~\ref{fig:ablation_core_components} (Right), using keyframe extracted from robot-view streams instead of egocentric view leads to 31.7\% drop in success rate.
% The VLM still captures the right moments (i.e., when the human points at each muffin), but a keyframe from the robot's view no longer identifies a single target because of \emph{parallax}: the same pointing gesture looks different depending on where it is viewed from.
% As shown in Figure~\ref{fig:example_figure_parallax}, in a keyframe from the robot's view, it is visually unclear which of the six muffins the finger is pointing at, so the keyframe does not specify the target for the low-level policy.
% In contrast, in a keyframe from the human's egocentric view, the finger and the target muffin line up along the line of sight, so the target is unambiguous.

\begin{figure}[h]
    \centering
    \includegraphics[width=0.3\textwidth]{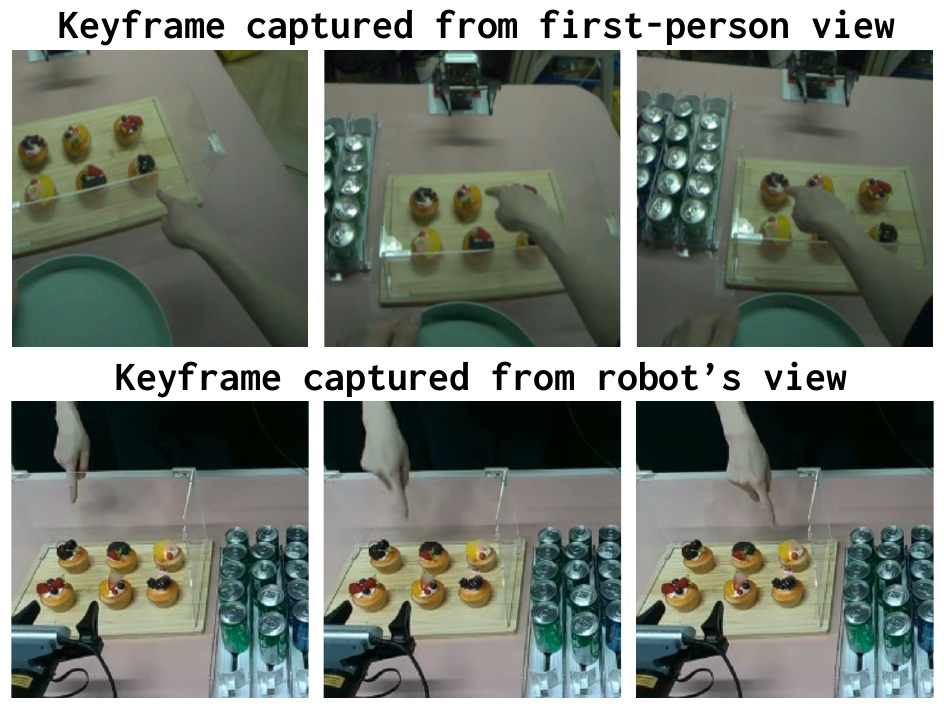}
    \caption{Example of the parallax. Keyframes captured from the human's first-person view clearly represents which object the human is pointing at. In contrast, referents of human's pointing gesture is hardly resolved in the keyframes captured from the robot's view due to parallax.}
    \label{fig:example_figure_parallax}
    \vspace{-0.3cm}
\end{figure}

\textbf{Ablation of egocentric view.}
Next, we examine the contribution of leveraging the egocentric context.
We compare \metabbr against a hierarchical baseline in which both policies operate on the robot-view stream instead: the high-level policy produces subtasks whose keyframes are captured from the robot's view, and the low-level policy is finetuned on the same segmented trajectories conditioned on these robot-view keyframes.
As shown in Figure~\ref{fig:ablation_core_components} (Right), this leads to a 41.7\%p drop in success rate in \texttt{Muffin-Serving}.
The VLM still captures the right moments (i.e., when the human points at each muffin), but a keyframe from the robot's view no longer identifies a single target because of \emph{parallax}: the same pointing gesture looks different depending on where it is viewed from.
As shown in Figure~\ref{fig:example_figure_parallax}, in a keyframe from the robot's view, it is visually unclear which of the six muffins the finger is pointing at, whereas in the egocentric view, the finger and the target muffin line up along the line of sight, so the target is unambiguous.

\begin{figure}[h]%{0.4\linewidth}
    \centering
    %\vspace{-0.6cm}
    \includegraphics[width=0.35\textwidth]{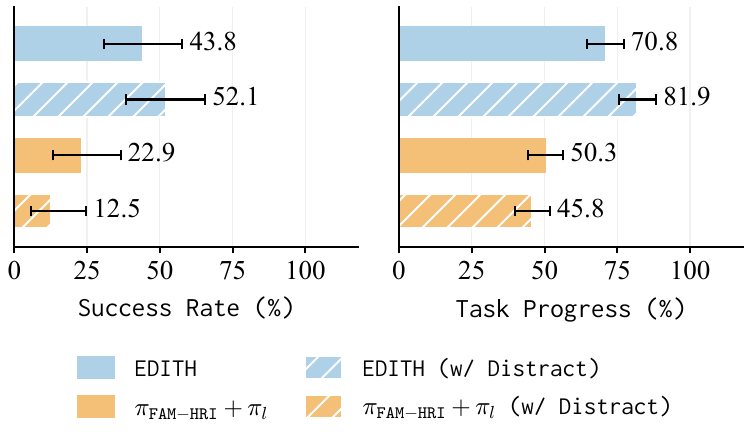}
    %\vspace{-0.7cm}
    \caption{Robustness of \metabbr to external distractions in the human's egocentric context (i.e., intermittently looking at a phone while requesting muffins), compared to $\pi_\texttt{FAM-HRI}+\pi_l$. While such distraction leads to degradation of $\pi_\texttt{FAM-HRI}+\pi_l$, \metabbr is unaffected by such distractions.}
    \label{fig:human_distraction}
\end{figure}

\textbf{Robustness of \metabbr to human distraction.}
When humans deliver their intent in natural interaction, their attention often shifts to things unrelated to the robot's task, such as 
% briefly 
checking a text on their phone.
We evaluate whether \metabbr remains robust under such distraction.
In the \texttt{Muffin-Serving} task, the human evaluator alternates between checking a text message and requesting a muffin, repeating this until
% all 
three muffins are requested.
As shown in Figure~\ref{fig:human_distraction}, the success rate of $\pi_\texttt{FAM-HRI}+\pi_l$
drops from 22.9\% to 12.5\%.
Since it selects the keyframe by a fixed rule on gaze fixation, it often picks a wrong frame when the human's gaze becomes noisy (e.g., while their eyes move between the phone and the muffin).
In contrast, \metabbr is unaffected by the distraction, achieving even higher success rate in the distracted setting.
This highlights the value of VLM-based keyframe extraction over a fixed rule on the gaze stream.
% This highlights the value of extracting keyframes with a VLM, rather than with a fixed rule on the gaze stream.

\subsection{Failure Analysis}
\label{subsec:failure}
\begin{figure}[h]
    \centering
    \includegraphics[width=0.3\textwidth]
    {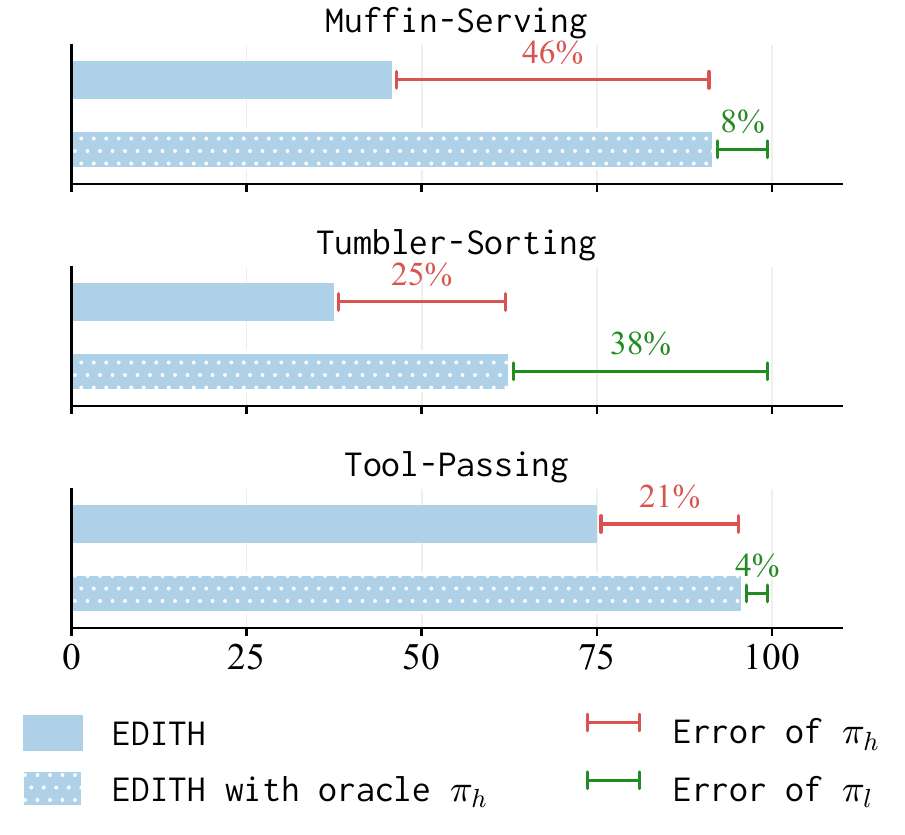}
    \caption{We disentangle the two sources of \metabbr's failure: an error of $\pi_h$ and an error of $\pi_l$. We visualize the former with \textcolor{red}{red} lines, while latter with \textcolor{forestgreen}{green} lines.}
    \label{fig:failure_analysis}                      % 아래쪽 여백 당김 (선택)
\end{figure}
In this section, we discuss the failure modes of \metabbr.
Failures come from two sources: errors of the high-level policy $\pi_h$ in producing subtasks and keyframes from the human's signals, and errors of the low-level policy $\pi_l$ in executing these subtasks.
To separate the two, we compare \metabbr with an oracle in which a human serves as $\pi_h$ and provides ground-truth subtask instructions and keyframes to $\pi_l$.
The gap between \metabbr and the oracle shows the failures caused by $\pi_h$, while the gap between the oracle and a 100\% success rate shows the failures caused by $\pi_l$ alone.
We conduct this analysis with two human evaluators.

As shown in Figure~\ref{fig:failure_analysis}, the main source of failure varies across tasks, but on average, $\pi_h$ accounts for a larger share of failures than $\pi_l$.
Qualitatively, most $\pi_h$ failures come from keyframes that are slightly off from the correct moment.
While the pretrained VLM extracts more accurate keyframes than the baselines, it is not trained on our data and thus still makes mistakes when the human makes requests in quick succession.
For $\pi_l$, failures occur even with correct keyframes and subtask instruction, for example when a keyframe shows a field of view not seen during training. In \texttt{Tumbler-Sorting}, large portion of $\pi_l$ failures are simple manipulation errors.
These results suggest two directions for improvement: training the VLM to extract more accurate keyframes, and collecting demonstrations from a more diverse set of humans to make $\pi_l$ more reliable.

\section{Conclusion}
\label{sec:conclusion}
We present \metabbr, a framework that lets robots follow human nonverbal signals by using the first-person view and gaze from smart glasses, processed in real time by a hierarchical policy. Experiments on three real-world interaction tasks show that \metabbr works well and stays robust to human distraction. We hope \metabbr is a step toward robots that interact with humans naturally and in real time.
%\dk{In this paper, we presented EDITH, a robot framework that understands human intent expressed through both verbal and nonverbal channels. Our key idea is to capture the human's first-person view and gaze as continuous streams via smart glasses and use them, together with language, as input to a hierarchical policy: a high-level policy infers the human's intent and decomposes it into subtask directives, while a low-level policy executes them on the robot at the control frequency. Across four realistic human-robot interaction tasks that require grounding visual attention, gestures, and ongoing activity, EDITH substantially outperforms baselines that rely on verbal instruction alone, and a user study further shows that it reduces the effort users spend conveying their intent, offering a more natural interface than language alone.}
%\section*{Acknowledgement}
%We thank Meta Reality Labs for providing Project Aria research kits used in this work, and Google Cloud for research credits through the Gemma Academic Program.

\section*{Acknowledgement}
We acknowledge the use of AI-generated images as illustrative elements in some of our figures.
They are used only for illustration purposes, and all qualitative results presented in this paper are from our real-world experiments.

\bibliographystyle{IEEEtran}
\bibliography{references}

\end{document}